\documentclass[runningheads]{llncs}
\usepackage[T1]{fontenc}
\usepackage{booktabs}
\usepackage{float}
\usepackage{hyperref}

\usepackage[utf8]{inputenc}
\usepackage{algorithm}
\usepackage[noend]{algpseudocode}
\usepackage{amsmath}

\usepackage{float}

\usepackage[belowskip=5pt]{caption} 

\usepackage{graphicx}
\usepackage{xcolor}
\usepackage{verbatim}
\begin{document}
\title{Safety Case Patterns for VLA-based driving
systems: Insights from SimLingo}
%
%
\author{Gerhard Yu\inst{1} \and
Fuyuki Ishikawa\inst{2} \and
Oluwafemi Odu\inst{1} \and Alvine B. Belle\inst{1}}
\authorrunning{G. Yu et al.}
%
\institute{York University, North York ON M3J 1P3, Canada \\
\email{\{gerhardy@, olufemi2@, alvine.belle@lassonde.\}yorku.ca}\\
\and
National Institute of Informatics, Chiyoda City Tokyo 101-8430, Japan\\
\email{f-ishikawa@nii.ac.jp}}

\maketitle              
\begin{abstract}
Vision–Language–Action (VLA)–based driving systems represent a significant paradigm shift in autonomous driving since, by combining traffic scene understanding, linguistic interpretation, and action generation, these systems enable more flexible, adaptive, and instruction-responsive driving behaviors. 
However, despite their growing adoption and potential to support socially responsible autonomous driving as well as understanding high-level human instructions, VLA-based driving systems may exhibit new types of hazardous behaviors. For instance, the integration of open-ended natural language inputs (e.g., user or navigation instructions) into the multimodal control loop may lead to unpredictable and unsafe behaviors that could endanger vehicle occupants and pedestrians. Hence, assuring the safety of these systems is crucial to help build trust in their operations.
To support this, we propose a novel safety case design approach called \textit{RAISE}. Our approach introduces novel patterns tailored to instruction-based driving systems such as VLA-based driving systems, an extension of Hazard Analysis and Risk Assessment (HARA) detailing safe scenarios and their outcomes, and a design technique to create the safety cases of VLA-based driving systems. A case study on \textit{SimLingo} illustrates how our approach can be used to construct rigorous, evidence-based safety claims for this emerging class of autonomous driving systems.

\keywords{	Requirement Engineering \and Safety case
\and Safety case patterns
\and Generative AI \and VLA-based driving system \and SimLingo.}
\end{abstract}
\section{Introduction}


System assurance is the process of justifying that mission-critical systems are dependable \cite{hawkins2016using}. Assurance cases can be used to support system assurance. They are a set of structured arguments that are supported by evidence and that allow demonstrating that a system sufficiently supports its intended non-functional requirements (e.g., safety, reliability, security) \cite{Maksimov2019}. 
Assurance case patterns are reusable templates derived from previously successful assurance cases, and aiming at supporting the development of new assurance cases \cite{oduautomatic}. 

VLA-based driving systems constitute an emerging class of ML-enabled autonomous driving systems. These systems enable minimal to no input from human drivers and still have the capability to reach their destination. These systems utilize several components to accomplish this. These include autonomous decision-making agents as well as sensors such as LiDAR, camera, and radar \cite{zhao2024autonomous}. The use of both agents and sensors helps autonomous vehicles perceive their surroundings, enabling them to avoid hazards. Thus, autonomous driving systems have started to be researched worldwide, especially in both industry and academia \cite{zhao2024autonomous}. Given the popularity and the safety-critical nature of these systems, ensuring their safety is crucial to avoid fatal accidents while also giving enough confidence in their practical use. 

The addition of human inputs or instruction-based control into safety-critical systems like VLA-based driving systems introduces significant safety hazards that are distinct from those encountered in traditional ML-enabled autonomous systems. For instance, a vehicle may follow a user’s instruction to reverse without adequately assessing the situation, potentially resulting in fatal collisions with pedestrians, other vehicles, or stationary obstacles. Similarly, user-initiated instructions to turn, accelerate, overtake, or react to complex scenarios such as approaching a yellow traffic light can lead to unsafe maneuvers, causing serious injuries, loss of life and substantial financial liabilities. Hence, the unpredictable nature of user-generated instructions can conflict with system defined actions for a given driving scenario, increasing the likelihood that the system will carry out hazardous actions in situations where it must reject unsafe instructions.

These hazards underscore the need for structured safety assurance when addressing VLA-based driving systems. The creation of safety cases for autonomous systems is resource-intensive and the resulting safety cases are difficult to maintain due to the increasing complexity and the heterogeneity of these systems  \cite{graydon2025examining}. For VLA-based systems, the challenge is further magnified since current methodologies lack structured patterns or reusable frameworks that specifically address the risks arising from using user instructions as input to these systems.

To address the aforementioned gaps, we make the following contributions to the system assurance field:

\begin{itemize}
    \item We propose an extension of HARA \cite{beckers2017structured} that allows analyzing traffic scenarios and driving actions to elicit the potentially dangerous actions that may be caused by user instructions, as well as the outcomes of the safe user instructions within appropriate operational scenarios. This allows linking instruction-level safety reasoning with event-based safety analysis.

    \item  We propose novel patterns that establish a foundational framework for demonstrating the safe behavior of VLA-based driving systems. These patterns capture the ability for these systems to evaluate user instructions and determine whether to accept appropriate instructions or reject potentially dangerous ones based on the operational context.
    
    \item We propose a novel safety case design approach called \textit{RAISE} (assuRance of vlA-based drIving SystEms) for constructing the safety cases of an emerging class of autonomous driving systems called VLA-based driving systems. \textit{RAISE} leverages the outcomes of the extended HARA and the proposed patterns to build concrete safety case models for these systems.

    \item We demonstrate the applicability of our approach through a case study that focuses on the creation of a safety case for \textit{SimLingo} \cite{renz2025simlingo}, a reference VLA-based driving system.
\end{itemize}

Our work is useful to researchers and practitioners, and more specifically to corporate safety analysts, system engineers, assurance case developers and regulatory authorities because it provides a structured framework for justifying the safe handling of user instructions in VLA-based driving systems.

\section{Background and Motivation}

\subsection{Advancement in VLA}
AI has rapidly expanded across many sectors, including healthcare, automotive, banking, and manufacturing, largely because it automates tasks that were once manual, repetitive, and time-consuming. This increasing popularity has been further accelerated by the development of Large Language Models (LLMs) which are built on transformer architectures and can understand and generate human-like text \cite{chang2024survey,hou2024large}. LLMs are capable of performing complex tasks such as translation, coding, and summarization \cite{hou2024large,oduautomatic}. 
Vision–language models extend LLMs' capabilities by combining visual and textual inputs, enabling tasks like image captioning and visual question answering. However, these two types of models are primarily limited to perception and interpretation and do not inherently possess the ability to execute real-world actions  \cite{sapkota2025vision,zhong2025survey}. VLA (Vision-Language-Action) models address these limitations by integrating perception, semantic reasoning, and action generation within a single multimodal framework \cite{zhang2025pure}. This unified design enables systems not only to understand visual and linguistic inputs but also to translate that understanding into task-oriented behaviors, thereby bridging the gap between multimodal perception and embodied interaction \cite{jiang2025survey}.

\subsection{VLA-based Driving Systems}

\begin{figure*}
\centering
\includegraphics[width=1\linewidth]{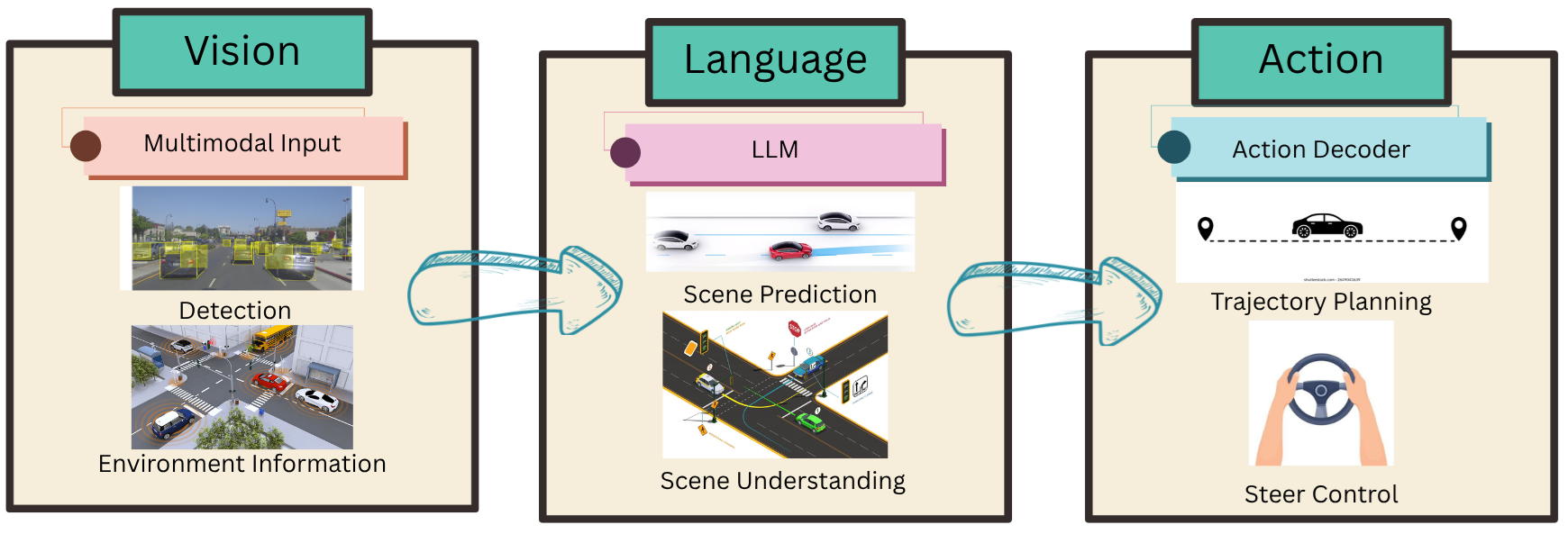}
\caption{Simplified architecture of VLA-based driving systems - adapted from \cite{jiang2025survey}} 
\label{fig:VLA-driving_systems_architecture}
\end{figure*}

VLA models, which can perceive environments, understand language, and follow instructions \cite{hu2025vision}, are increasingly used in autonomous driving \cite{jiang2025survey}. They help vehicles interpret surroundings, navigate safely, and explain their decisions and planned movements in natural language \cite{renz2025simlingo,shao2024lmdrive}. This transparency helps passengers understand what the vehicle is doing and increases their sense of safety. As a result, several VLA-based driving systems such as \textit{SimLingo}  \cite{renz2025simlingo}  and \textit{LMDrive}  \cite{shao2024lmdrive} have been developed. As shown in Figure \ref{fig:VLA-driving_systems_architecture}, VLA-based driving systems use multimodal inputs and produce outputs that include both driving actions and textual explanations.

The CARLA Leaderboard \cite{carla_leaderboard_2020,carla_leaderboard_2023} is a very popular autonomous driving simulator used to develop, train, and test autonomous driving systems \cite{dosovitskiy2017carla}. It evaluates them by running simulations on predefined routes with different traffic situations, locations, and weather conditions \cite{dosovitskiy2017carla}. Results from this leaderboard show that VLA-based driving systems perform well, indicating they have the potential to autonomously drive passengers to their desired destinations.

\vspace{-3em} 

\subsection{Existing Assurance Effort}

An assurance case is a structured, reasoned, and auditable collection of arguments, supported by evidence, that demonstrates a system’s non-functional requirements (e.g., safety and security) have been correctly implemented \cite{delavara2017,Mohamad2021}. Several studies (e.g., \cite{denney_pai_2013,gleirscher2017,carlan2020,annable2025,zeroual_hamid2023,zeroual_july_2023,habli2025big,hawkins2021guidance,selviandro2021assurance,bandur2023}) focused on the creation of argument patterns and their use to create assurance cases (e.g., safety cases, security cases). Most of these studies emphasize the reuse of argument structures to improve assurance case quality, reduce argument development effort, and support the systematic construction of assurance arguments across similar systems and/or application domains. Gauerhof \cite{Gauerhof2020} relied on previous work \cite{picardi2020} to assure the safety of Machine Learning for pedestrian detection at crossings. 

More recent system assurance methodologies for ML-enabled systems in general and ML-enabled autonomous systems in particular include the BIG-Argument \cite{habli2025big}, PRAISE (PRinciples-bAsed EthIcs assurance) \cite{porter2024}, AMLAS \cite{hawkins2021guidance}, and AMLAS-r \cite{hu2025arguing}. These methodologies often provide structured argumentation processes, argument patterns, and reliability-focused extensions to justify the dependability (e.g., safety) of ML components. Often combined with hazard analysis techniques such as HARA to identify and mitigate risks, some of these methods have been applied to the assurance of various ML-enabled driving systems. These include the ML-enabled trajectory prediction component of an open-source autonomous driving system called Baidu Apollo \cite{odu2025llm}, the Reinforcement Learning components used by Quanser autonomous vehicles \cite{sivakumar_quanser_2024}, and  SMIRK (an ML-enabled pedestrian automatic emergency braking system) \cite{borg2023ergo,socha2022smirk}. 

Furthermore, some approaches (e.g., \cite{viger2024ai,gohar2024codefeater,oduautomatic,sivakumar2024prompting,odu2025smartgsn,shahandashti2024using}) have either explored the automatic generation of assurance cases or their defeaters (i.e., deficits) using LLMs or integrated safety analyses like STAMP/STPA to assure level 4 autonomous vehicles (e.g., \cite{kodama2024case}). Our work is closer to existing approaches (e.g., \cite{sivakumar_quanser_2024,kodama2024case}) that focused on manually creating safety cases for ML-enabled autonomous driving systems. However, such approaches are not suitable to demonstrate the safety of VLA-based driving systems because they do not address the emerging types of hazards arising from the reliance on user instructions to drive vehicles. This significantly reduces their applicability and replicability when it comes to assuring the safety of these newer instruction-driven vision AI models.

\section{Proposed Method}

\subsection{Overview} 

We propose a novel approach called \textit{RAISE} that enables safety professionals to systematically, rigorously, and consistently develop safety cases for an emerging class of systems known as instruction-based driving systems. Our novel approach for constructing safety assurance cases is particularly suitable for VLA–based autonomous driving systems. 
\textit{RAISE}  begins with the execution of an extended version of HARA, which consists of hazard identification, risk analysis, risk evaluation, risk control, and safe event identification. It then proceeds with the definition of a library of GSN-compliant argument patterns suitable for VLA-based driving systems and addressing two key safety objectives inherent to VLA-based driving systems: rejecting dangerous user instructions and executing instructions that are safe. \textit{RAISE} also comprises an algorithm that allows creating safety cases for VLA-based driving systems and that we derived from the HARA process outcomes and the library of GSN-compliant patterns. 

\begin{figure}
\centering
\includegraphics[width=1\linewidth]{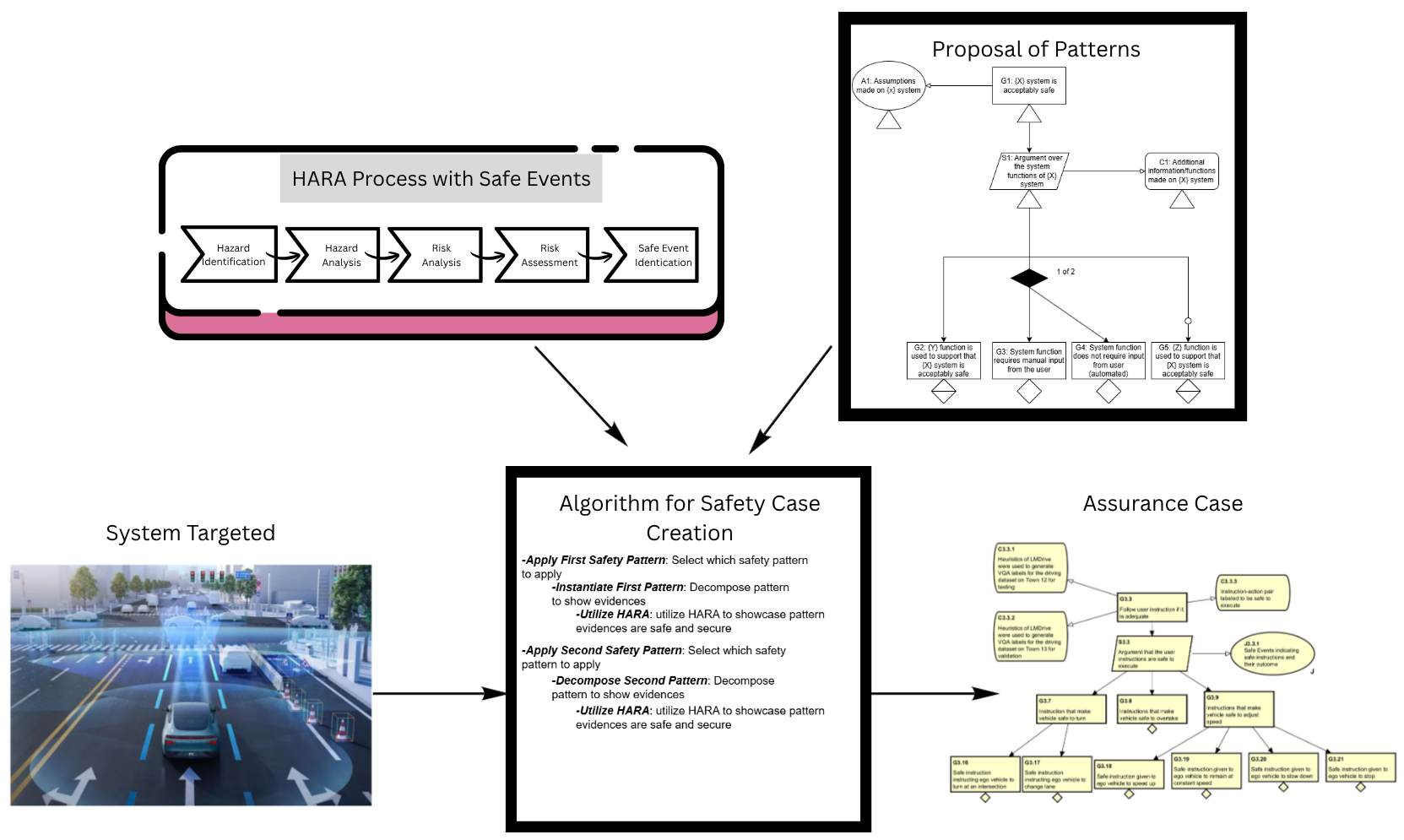}
\caption{High-level Overview of the \textit{RAISE} approach
}
\label{fig:raise_appproach_overview}
\end{figure}

To create the safety case of a given system,  our algorithm iteratively selects a suitable pattern from the library to instantiate each portion of the safety case, and if needed, iteratively decomposes the safety case goals according to key operational criteria (e.g.,  hazards, scenarios, or modes) and links each decomposition step to specific HARA operational scenarios. This decomposition allows for progressively generating child goals until they can be supported by concrete evidence from the VLA-based driving system under analysis. The output of this algorithm is a concrete safety case in which the top-level goal argues that the VLA-based driving system at hand is sufficiently safe.
Figure \ref{fig:raise_appproach_overview} depicts the high-level overview of the \textit{RAISE} approach. We further describe it in the remainder of this Section.

\subsection{Patterns for Instructions}

Using the HARA process described above, we have created the safety case of reference VLA-based driving system called \textit{SimLingo}. By using HARA, we were able to identify the main goals of \textit{SimLingo}. Table \ref{tab:System Functions} reports them. We then proceeded to extract common scenarios where a vehicle might find itself and use those as operational scenarios. We then use that knowledge to help decompose the top goals of \textit{SimLingo}, which leads to our current top-level safety goal as shown in Figure \ref{fig:top-argument_simlingo}. To generalize this to other VLA-based driving systems, we also analyzed the specific scenarios that we have gathered from other VLA-based driving systems such as \textit{LMDrive}. Thus, we used the insights from this analysis to create two distinct patterns that showcase some common scenarios VLA-based driving systems may encounter and what their resulting decomposition would look like depending on the context of the user instruction. We refer to our two patterns as the \textit{RI (Reject Instruction)} and \textit{AAI (Accept Adequate Instructions)} patterns, respectively. These patterns are part of the library of patterns and contain \textit{hot spots}, i.e., placeholders filled with generic (i.e., abstract) information \cite{carlan2020enhancing}. 
When instantiating these patterns, safety analysts can replace these \textit{hot spots} with VLA-driving system-specific information. These patterns are also meant to be extended. For instance, if safety analysts identify \textit{cold spots} (i.e., potential deficits) in these patterns, they can further refine existing patterns by fixing them.

\begin{figure}
\centering
\includegraphics[width=1\textwidth]{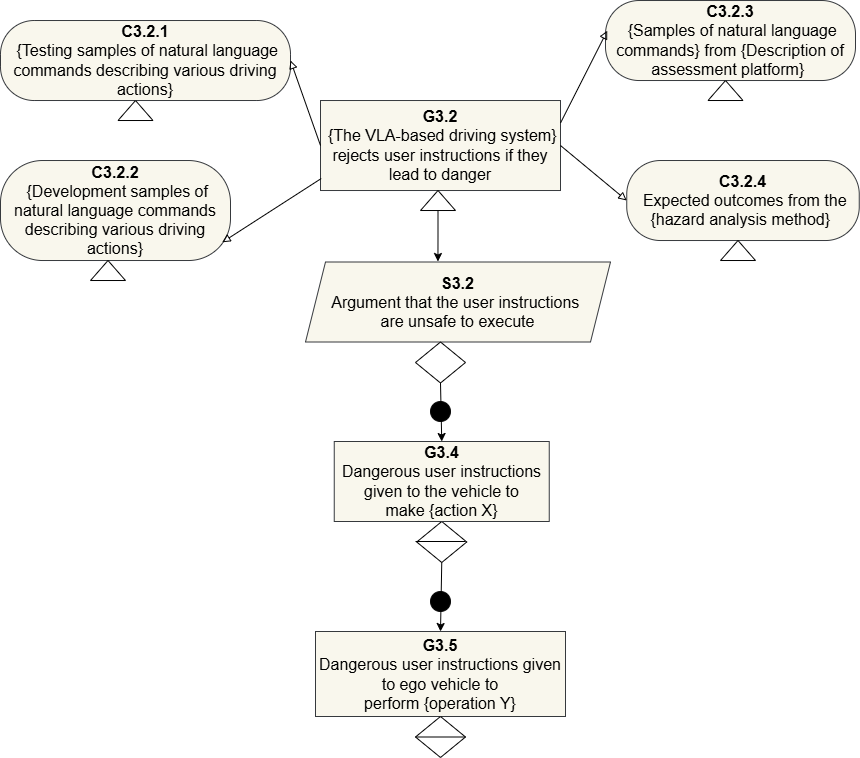}
\caption{Reject Instruction Pattern represented in the GSN}
\label{fig:reject_pattern}
\end{figure}

\begin{figure}
\centering
\includegraphics[width=1\textwidth]{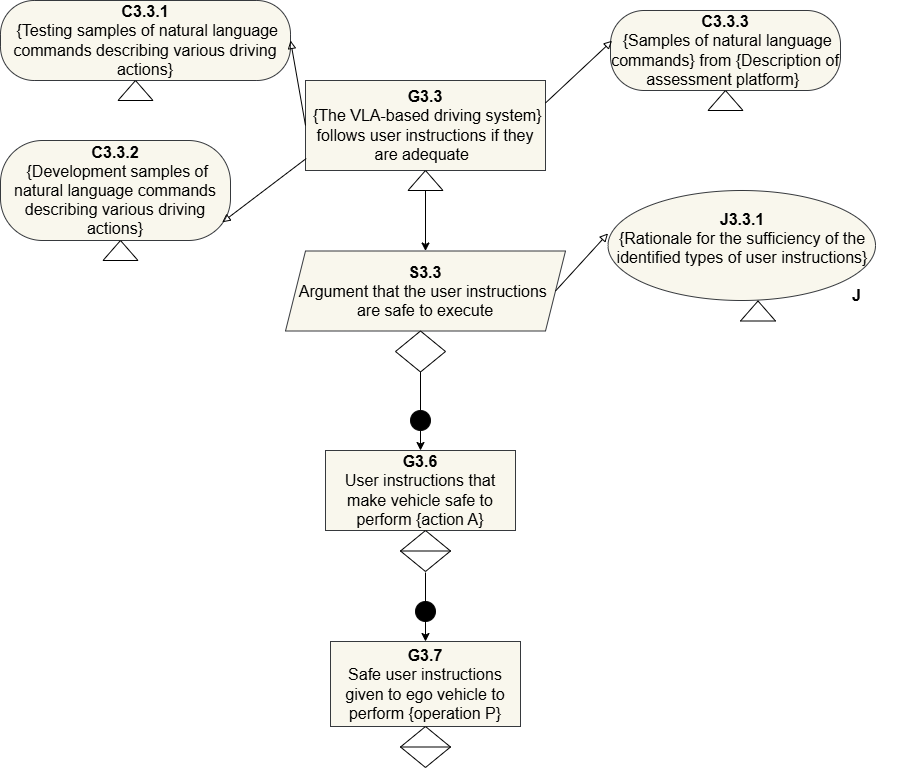}
\caption{Accept Adequate Instructions Pattern represented in the GSN} 
\label{fig:accept_pattern}
\end{figure}

Figure \ref{fig:reject_pattern} describes the \textit{RI} pattern. This pattern allows arguing the safety surrounding the capability of a given VLA-based driving system to reject a dangerous instruction. This pattern showcases common scenarios that the vehicle might find itself in when trying to safely handle user instructions, such as the vehicle being at an intersection in which a dangerous instruction could lead to a crash. Figure \ref{fig:accept_pattern} represents the \textit{AAI} pattern.  This pattern allows arguing the safety surrounding the capability for a given VLA-based driving system to accept safe and/or adequate user instructions. 

Using both patterns and insights from the HARA process, we then introduce a generic approach to create safety cases for instruction-based VLA systems. Algorithm \ref{algorithm:raise}
 describes this approach. This algorithm operationalizes our new safety case design approach. This algorithm provides professionals with some guidance on how to create a safety case for VLA-based driving systems. This algorithm aims to systematically construct a safety case for a VLA-based driving system by first analyzing system documentation and outputs. Then, it applies an appropriate hazard analysis and/or risk assessment technique (e.g., HARA) to identify hazards and safety goals. Using the results of the risk assessment technique, it then relies on patterns to incrementally build a structured argument, starting from a top safety goal, and iteratively decomposing it into strategies, sub-goals, and supporting evidence.

\begin{algorithm} 
\caption{: RAISE}
\label{algorithm:raise}
\begin{algorithmic}[1]
 \Statex \textbf{Inputs:} analyzed VLA-based driving system; specific driving scenarios of a set of VLA-based driving systems (e.g., analyzed system); list of risk of hazard analysis and/or risk assessment techniques (e.g., HARA, FMEA, HAZOP) 

 \Statex \textbf{BEGIN}

\State $\textit{System\_Docs} \gets
\textsc{GetArtifacts}(\textit{Analyzed\_VLA\_System})$

\State $\textit{Technique} \gets \textsc{SelectRiskAssessmentTechnique}(\textit{TechniqueList})$

\State $\textit{Risk\_Results} \gets
\textsc{RiskAssessment}(\textit{Technique})$

\State $\textit{Pattern\_Library} \gets \textsc{CreatePatternLibrary}(\textit{VLA\_System\_Set})$

\State $\textit{Safety\_Case} \gets EmptyGoalStructure()$

\State $\textit{Safety\_Case}+ \gets
\textsc{CreateTopLevelArgument}(\textit{Risk\_Results})$

\ForAll{$\textit{pattern}  ~in~  \textit{Pattern\_Library}$}
    \State $\textit{Safety\_Case}+ \gets \textsc{InstantiatePattern}(\textit{pattern, Risk\_Results})$
    
     \State $
    \textsc{RefineSafetyCase}(\textit{Safety\_Case})$
    
\EndFor

\ForAll{$\textit{leaf\_goal} 
 ~in~   \textit{Safety\_Case}$}
    \State $\textit{solution} \gets
    \textsc{FindSuitableSolution}(\textit{System\_Docs})$

    \State $\textit{Safety\_Case}+ \gets 
     \textsc{CreateSupportedBy}(\textit{leaf\_goal}, \textit{solution})$
     
      \State $\textit{Safety\_Case}+ \gets 
     \textit{solution}$
\EndFor

\State \Return $\textit{Safety\_Case}$

\Statex \textbf{END}
\end{algorithmic}
\end{algorithm}

\subsection{HARA in our context}

We performed HARA \cite{beckers2017structured} on \textit{SimLingo} to help determine the criticality of this system and develop safety goals for \textit{SimLingo}  \cite{johansson2015importance}, identifying hazards that are present in the system and addressing them. Following the HARA process, we first determine the definition of \textit{SimLingo}. Once we have identified the definition of \textit{SimLingo}, we then proceed with the creation of assumptions for this system, one of these assumptions being: \textit{SimLingo} is in action mode. Once we finish creating the assumptions of the system, we then determine the system functions of \textit{SimLingo} as indicated in Table \ref{tab:System Functions}. These system functions give us an insight into the primary objectives of what \textit{SimLingo} is trying to achieve. These systems' function, combined with the operational scenarios in Table \ref{tab:Operational scenarios}, gives us insight into dangerous instructions on those operational scenarios and their effect. Upon finishing the definition of the system functions of \textit{SimLingo}, we then proceeded to list the malfunctions of each system function, which gives us insight as to how those system functions may fail. 

\begin{table*}
    \centering
    \caption{HARA: System Functions}
    \begin{tabular}{|p{1.1cm}|p{10.5cm}|}
    \hline
        \textbf{ID} & \textbf{System Function} \\
        \hline
         SF1 & \textit{SimLingo} is able to process information properly based from the camera\\
         \hline
         SF2 &  \textit{SimLingo} is able  to reach the specified destination \\
         \hline
         SF3 &  \textit{SimLingo} is able  to provide commentary based from the current driving scenario \\
         \hline
         SF4 &  \textit{SimLingo} is able  to align the action space \\
         \hline
         SF5 &  \textit{SimLingo} is able  to reject dangerous instructions \\
         \hline
    \end{tabular}
    \label{tab:System Functions}
\end{table*}

\begin{table}
    \centering
    \caption{HARA: Operational Scenarios}
    \begin{tabular}{|p{1.5cm}|p{5cm}|}
    \hline
        \textbf{ID} & \textbf{System Function} \\
        \hline
         OS1 & Vehicle at an intersection\\
         \hline
         OS2 & Vehicle changing lanes \\
         \hline
         OS3 & Vehicle u-turning \\
         \hline
         OS4 & Vehicle overtaking \\
         \hline
         OS5 & Vehicle increasing speed \\
         \hline
         OS6 & Vehicle remaining at constant speed \\
         \hline
         OS7 & Vehicle slowing down \\
         \hline
         OS8 & Vehicle stopping \\
         \hline
         OS9 & Vehicle reversing \\
         \hline
    \end{tabular}
    \label{tab:Operational scenarios}
\end{table}

Further conducting HARA, we then determine some of the operational scenarios of \textit{SimLingo}, as indicated in Table \ref{tab:Operational scenarios}. These operational scenarios are scenarios that we consider for what a vehicle will normally do when driving. These operational scenarios give us an insight into what the effects of dangerous instructions will have on those scenarios. Upon finishing the operational scenarios, we then proceeded to \textit{Analyze the hazardous events}. This process determines the outcome/effect of each operational scenario when the system function malfunctions, along with a risk assessment indicating how severe, how likely, and how controllable the hazard is, and, lastly, a safety goal connected to each hazard deemed to be of top priority.

Recall from above that we have further extended HARA to include \textit{Safe Events}, which showcase the effects that safe instructions have on those operational scenarios. The \textit{Safe Events} do not include the risk assessment portion of \textit{Analyze the hazardous events} along with their appropriate safety goals, since this portion itself inherently focuses on non-hazardous instructions and scenarios. This extension is used to identify the expected outcomes when \textit{SimLingo} accepts safe instructions. We also use this outcome to create our second pattern. Overall, we used HARA to gain an understanding of how the overarching goals of user-instructions, that is usually context-based, can be further decomposed to give a clearer and well-formed safety case. Our complete HARA results are available on GitHub\footnote{Link to the results of the HARA process: \href{https://anonymous.4open.science/r/RAISE-repository-C3D8/}{https://anonymous.4open.science/r/RAISE-repository-C3D8/} 
}.

\subsection{Creation of Concrete Assurance Goals}

To create our safety case, we followed the design steps outlined in our algorithm. We represented the safety case using the Goal Structuring Notation (GSN), a widely used graphical notation \cite{kelly2004goal,wei2019model,shahandashti2024prisma,Belle2023,sivakumar_quanser_2024,trentinaglia2022}. Using HARA, we have determined the main safety goals of \textit{SimLingo} through the system functions. These system functions form the overarching goals of our safety case. An example of this is one of the system functions of \textit{SimLingo}, which is SF1. The latter indicates that \textit{SimLingo} is able to process information properly based on the camera, which appears as one of the goals in our safety case. This indicates that the camera is a key component for the vehicle to be able to perform properly in the simulation.

Another example is through the system function of SF5, which suggests that \textit{SimLingo} is able to reject dangerous instructions. This system function forms the rejecting instruction goal for SF5. While the system functions do not indicate accepting safe instructions, to reach the specified destination, the vehicle should still be able to accept safe instructions to prevent accruing too many penalty points, thus forming the safety goal of \textit{SimLingo} being able to accept safe instructions. Once identified, we then focused on the instruction aspect of \textit{SimLingo}, where the operational scenarios come from. These operational scenarios then allow creating the subsequent child goals of \textit{SimLingo}, which indicate it is able to accept safe instructions and reject dangerous instructions. These child goals obtained from the HARA results serve as a basis for identifying outstanding safety requirements, thereby informing safety professionals of any remaining safety deficits that must be addressed to achieve an acceptable level of system safety.

\begin{figure}
\centering
\includegraphics[width=0.9\linewidth]{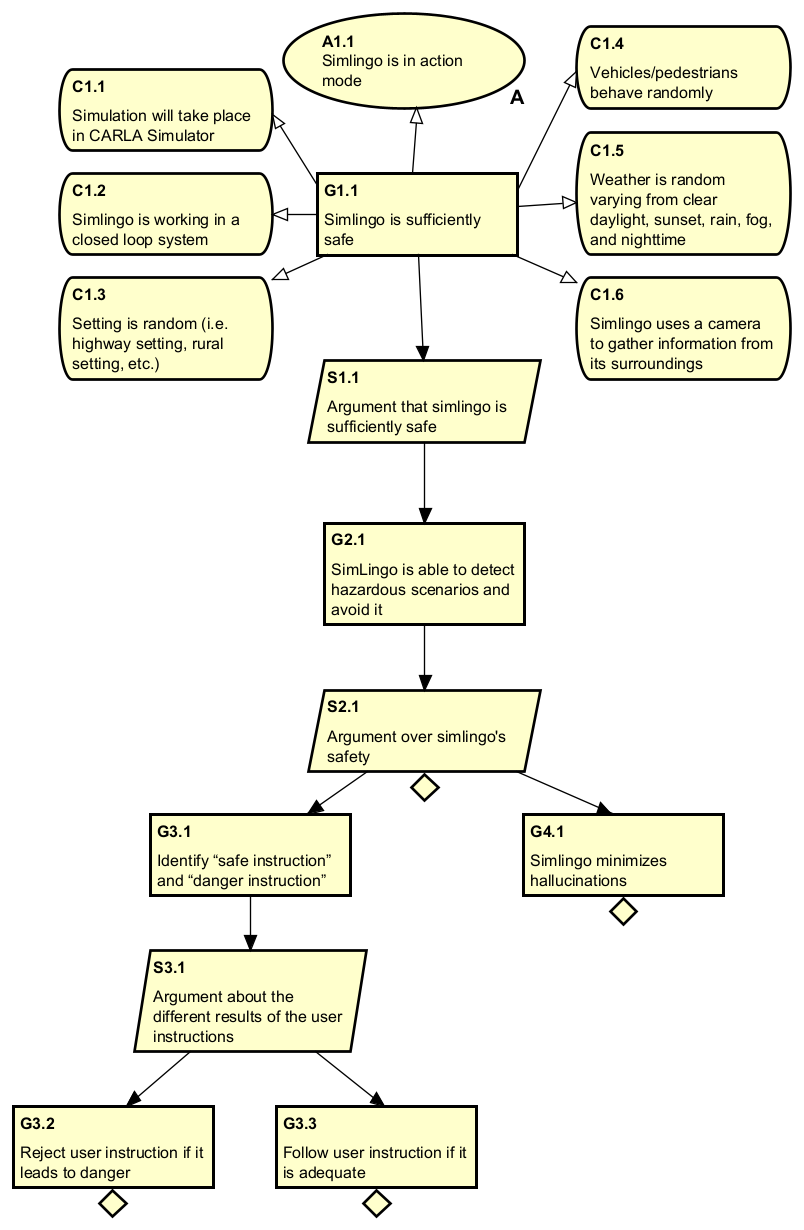}
\caption{Excerpt of the Top-Level Argument Structure of the \textit{SimLingo}  \cite{renz2025simlingo} Safety case
} 
\label{fig:top-argument_simlingo}
\end{figure}

Figure \ref{fig:top-argument_simlingo} depicts some of the GSN elements comprised in the top-level argument of the \textit{SimLingo} safety case. In Figure \ref{fig:top-argument_simlingo}, G1.1 is the top goal. This goal aims at demonstrating that \textit{SimLingo} is sufficiently safe to use. G1.1 is connected to several contexts. These include: C1.1, which states that \textit{SimLingo} execution takes place in the CARLA simulator; C1.2, which states that \textit{SimLingo} is a closed loop system; 
C1.4, which states that vehicles/pedestrians behave randomly; 
and C1.6, which states that \textit{SimLingo} uses a camera as its sensor. These contexts respectively show under which conditions \textit{SimLingo} is operating and also specify its settings. G1 relies on a strategy called S1.1. The latter explains how to decompose G1.1 into sub-goals to show that \textit{SimLingo} is indeed safe. Furthermore, goals such as G3.2 and G3.3 respectively demonstrate that \textit{SimLingo} is able to reject dangerous user instructions and accept safe user instructions.

\begin{figure}
\centering
\includegraphics[width=1\linewidth]{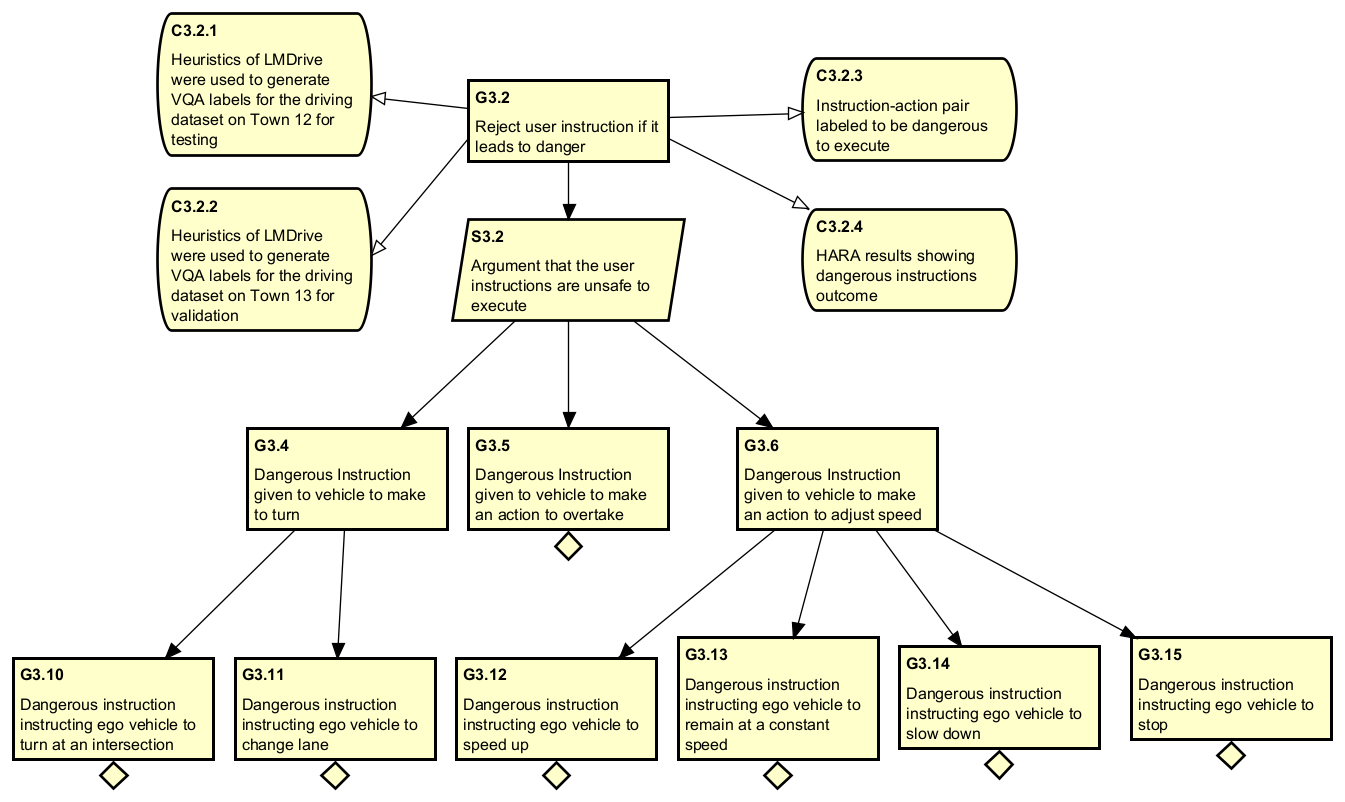}
\caption{Partial Safety Case of \textit{SimLingo} \cite{renz2025simlingo} obtained from the instantiation of the reject instructions pattern} 
\label{fig:reject_pattern_instantiation}
\end{figure}

\begin{figure}
\centering
\includegraphics[width=1\linewidth]{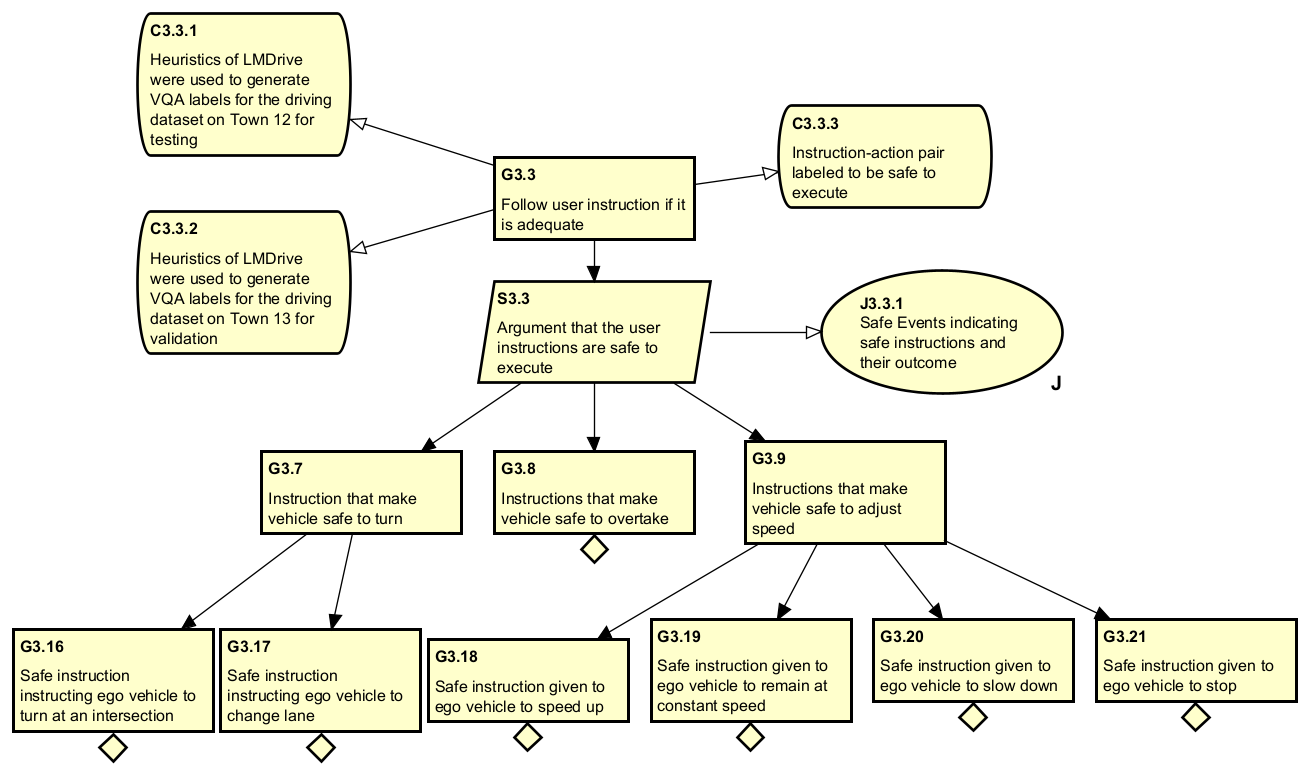}
\caption{Partial Safety Case of \textit{SimLingo} \cite{renz2025simlingo} obtained from the instantiation of the accept adequate instructions pattern} 
\label{fig:accept_pattern_instantiation}
\end{figure}

To further develop the argumentation structure of the \textit{SimLingo} safety case, we respectively instantiated the \textit{reject instructions} pattern (see Figure  \ref{fig:reject_pattern}) and the \textit{accept adequate instructions} pattern (see Figure  \ref{fig:accept_pattern}). This allowed us to design the argumentation structures of G3.2 and G3.3. 
Figure \ref{fig:reject_pattern_instantiation} and Figure \ref{fig:accept_pattern_instantiation} respectively show the resulting argumentation structures of both goals.

\section{Discussion}
The \textit{RAISE} approach enables the systematic identification, mitigation, and documentation of potential safety risks and safe events induced by user instructions. It also provides a framework to help safety case developers create concrete safety case models that sufficiently justify the safety of instruction-based driving systems. We believe focusing on the assurance of such systems is crucial because these instructions induce new safety risks. Systematically addressing these risks can strengthen confidence in the development, deployment and use of VLA-based driving technologies.

Three main insights emerge from our proposed approach and our case study. The first insight is that the safety of instruction-based VLA systems revolves around several objectives including the rejection of dangerous instructions and the acceptance of safe instructions. The second insight is that instruction safety is very context-dependent, requiring instructions to explicitly be labeled as safe or dangerous. The third insight is that HARA traditionally only deals with hazardous situations; thus, we have extended HARA to support the notion of \textit{Safe Events}, which allowed us to elicit safe instructions and their outcomes. These insights, together with our \textit{RAISE} approach, can provide clear guidance to the safety engineers of several automotive companies, such as Toyota, Tesla, and Waymo, in developing robust safety cases for their instruction-based autonomous driving systems.

Although VLA-based autonomous driving systems are relatively new, we believe that by raising concerns regarding the new types of unsafe behaviours these systems may exhibit and by proposing an initial approach to assure their safety, we will encourage researchers and professionals to further explore this topic. This will help develop more trustworthy VLA-based systems and facilitate their compliance with industry standards.

\section{Future Plans}

We will develop our ideas into a full-length paper by extending \textit{RAISE} into a comprehensive safety assurance framework for VLA-based driving systems. Hence, we will extend our library of GSN-compliant argument patterns to cover additional safety objectives, including robustness to ambiguous or adversarial language instructions, multimodal perception inconsistencies, uncertainty handling, explainability, and ethical decision-making in socially interactive traffic scenarios. We will also refine our \textit{RAISE} algorithm to support traceability and adaptability to evolving operational scenarios. Finally, we will conduct additional case studies using diverse VLA-based driving systems (e.g., AutoVLA \cite{Zhou_Cai_2026}, OpenDriveVLA \cite{Zhou_Han_2026}, EMMA \cite{Hwang2024}, LangCoop \cite{Gao_Langcoop2025}, DiffVLA \cite{Jiang_DiffVLA2025}) across urban, highway, and mixed-traffic environments to further evaluate \textit{RAISE}’s generalization and effectiveness for evidence-based safety assurance.

\begin{credits}
\subsubsection{\ackname} This work was undertaken thanks in part to funding from the Connected Minds program, supported by Canada First Research Excellence Fund, grant \#CFREF-2022-00010.

\subsubsection{\discintname}
The authors have no competing interests to declare that are relevant to the content of this article. 
\end{credits}
%
%
%
\bibliographystyle{splncs04}

\end{document}